\documentclass[letterpaper, 10 pt, conference]{ieeeconf}  

\IEEEoverridecommandlockouts                              

\usepackage{amsmath} 
\usepackage{amssymb}  
\usepackage{mathrsfs}
\usepackage{graphicx} 
\usepackage{epsfig} 
\usepackage{xcolor}
\usepackage{bm}
\usepackage{cite}
\usepackage{balance}

\newtheorem{theorem}{\textbf{Theorem}}

\newtheorem{definition}{\textbf{Definition}}

\newtheorem{remark}{\textbf{Remark}}

\newcommand{\col}{\textrm{col}}

\newcommand{\diag}{\textrm{diag}}

\newcommand{\op}{\textrm{op}}
\newcommand{\des}{\textrm{des}}

\newcommand{\gap}{\vspace{0.1cm}}
\DeclareMathOperator*{\argmin}{argmin}

\begin{document}


\title{\LARGE \bf
Safety-Critical Coordination of Legged Robots via Layered Controllers  \\ and Forward Reachable Set based Control Barrier Functions
}


\author{Jeeseop~Kim$^{1}$, Jaemin~Lee$^{1}$, and Aaron~D.~Ames$^{1}$
\thanks{This work is supported by Dow under project  \#227027AT and TII X1 under project \#A6847}
\thanks{$^{1}$J. Kim, J. Lee, and A. D. Ames are with the Department of Mechanical and Civil Engineering, California Institute of Technology, Pasadena, CA 91125, USA, {\tt\small \{jeeseop, jaemin87, ames\}@caltech.edu}}%
}

\maketitle

\begin{abstract}

This paper presents a safety-critical approach to the coordination of robots in dynamic environments. To this end, we leverage control barrier functions (CBFs) with the forward reachable set to guarantee the safe coordination of the robots while preserving a desired trajectory via a layered controller.
The top-level planner generates a safety-ensured trajectory for each agent, accounting for the dynamic constraints in the environment.
This planner leverages high-order CBFs based on the forward reachable set to ensure safety-critical coordination control, i.e., guarantee the safe coordination of the robots during locomotion.
The middle-level trajectory planner employs single rigid body (SRB) dynamics to generate optimal ground reaction forces (GRFs) to track the safety-ensured trajectories from the top-level planner.
The whole-body motions to adhere to the optimal GRFs while ensuring the friction cone condition at the end of each stance leg are generated from the low-level controller. 
The effectiveness of the approach is demonstrated through simulation and hardware experiments.

\end{abstract}

\section{Introduction}
\begin{figure}[t!]
\centering
\includegraphics[draft=false, width=\linewidth]{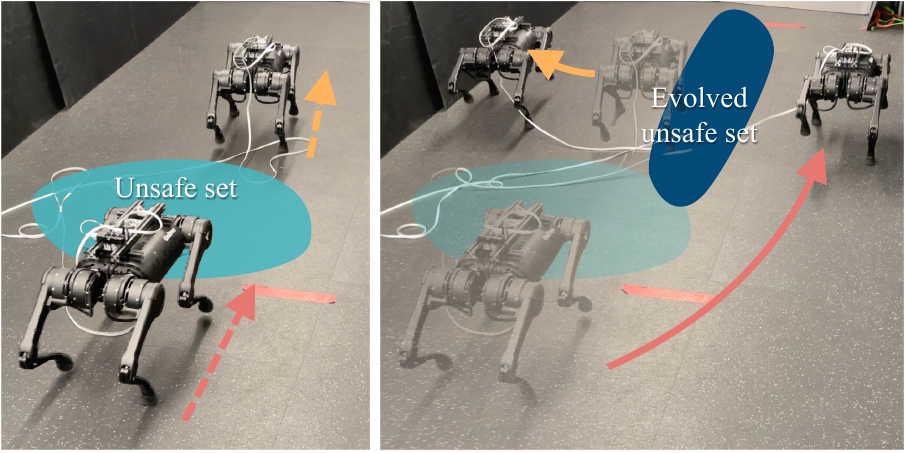}
\vspace{-2.0em}
\caption{Snapshots illustrating the safety-ensured coordination with forward reachable set based CBF in the environment with moving objects.}
\label{fig:firstpage}
\vspace{-2.0em}
\end{figure}

Safety-ensured coordination of the legged robots operating in dynamic environments which include moving obstacles is an essential component in enabling the cooperative work between robots, and assisting people in human-centered environments, e.g., performing collaborative tasks in urban settings.
One of the essential problems in deploying legged robots in dynamic environments is the mitigation of potentially conflicting objectives: the robots need to track the desired trajectory while maneuvering safely to avoid approaching objects (see Fig. \ref{fig:firstpage}).
While the coordinated control of multi-robot systems (MRSs) in static/dynamic environments has been studied in the context of 
ground and aerial vehicles \cite{fiorini1998motion,schneider2003potential,machado2016multi, tagliabue2019robust, wehbeh2020distributed},
the control of legged multi-robots in dynamic environment presents new and unique challenges.
The coordinated locomotion of the legged robots in dynamic environments (i.e., environments with moving obstacles) is inherently high-dimensional, highly dynamic, and hybrid in nature, adding complexity to the synthesis of safety-ensured coordination controllers that maintain stable locomotion. 

\subsection{Related Work}
Several studies have introduced control methodologies for stable and robust coopeƒrative locomotion of a team of legged robots. These frameworks have utilized a variety of techniques: the linear inverted pendulum (LIP) model \cite{kim2022cooperative}, single rigid body (SRB) dynamics \cite{kim2022layered}, and data-driven trajectory planners \cite{fawcett2022distributed}. Furthermore, safety-critical collaborative legged locomotion has been studied with static obstacles \cite{kim2023safety}. However, safe coordination of legged robots in dynamic environments has not been studied actively. 

Ensuring safety among moving objects is essential in the coordination control of the robotic systems. 
In the domain of motion planning and collision avoidance, 
potential field methods \cite{song2002potential,zhang2010dynamic} and velocity obstacles \cite{van2011reciprocal, guo2021vr} have been studied on various platforms but not widely on legged robots.
\textit{Control barrier functions} (CBFs) \cite{ames2016control, ames2019control} have emerged as a popular tool for achieving safety guarantees on robotic systems. In the context of legged robots, recent studies have introduced CBF-based planners/controllers to generate safe trajectory via a layered architecture \cite{molnar2021model, grandia2021multi, lee2023hierarchical} and ensure safe whole-body motion control \cite{khazoom2022humanoid}.
In the setting of MRSs composed of ground and aerial vehicles, CBFs have been successfully applied to achieve collision-free behavior \cite{wang2017safety, zhao2017defend, pickem2017robotarium, chen2020guaranteed}. 
However, designing CBFs and implementing corresponding safe control inputs typically leverage the distance from the unsafe set, resulting in robots avoiding the obstacles only when in close proximity.
To improve upon this,  Model predictive control approaches \cite{zhu2019chance, lindqvist2020nonlinear} with CBFs \cite{zeng2021safety} and HJ reachability analysis \cite{mitchell2005time, bansal2017hamilton, lee2020reachability} have demonstrated the effectiveness of guaranteeing safety in dynamic environments with ground and aerial vehicles.
However, addressing the \textit{safety} in the \textit{coordination of the legged multi-robot systems} in the \emph{dynamic environments with moving objects} has not been studied actively.






\subsection{Contributions}
To address the safety-critical coordination of the legged robots in dynamic environments, this paper makes three key contributions: 1) a safety-critical CBF-based coordination planner that can guarantee the safety of legged multi-robot systems, 2) a control framework that effectively addresses locomotion while tracking the optimal trajectories for safe coordination, and 3) experimental evaluation on a pair of quadrupeds.

\begin{figure}[t!]
\centering
\includegraphics[draft=false, width=\linewidth]{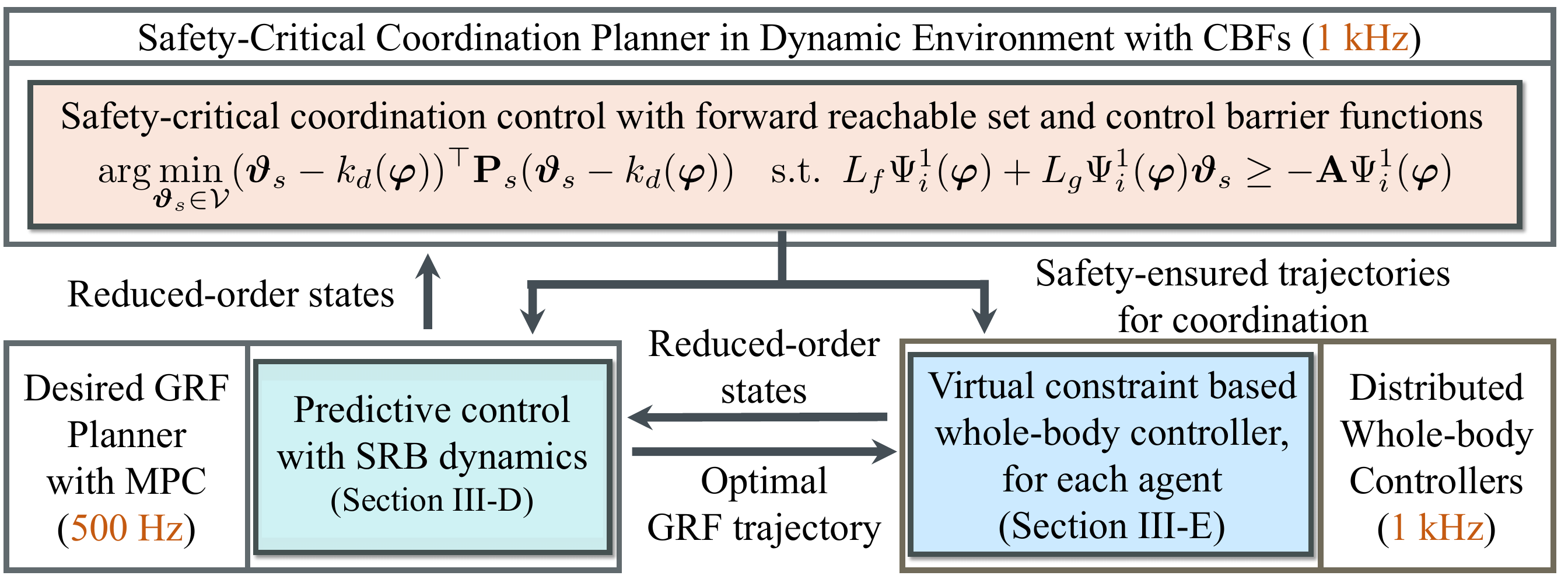}
\vspace{-1.5em}
\caption{Overview of the proposed control architecture with safety-critical coordination controller at the top, desired GRF planner with MPC in the middle, and the distributed whole-body controller in the bottom layer.}
\label{fig:architecture}
\vspace{-1.5em}
\end{figure}

To realize the key contributions of the paper and thereby ensure safety-critical coordinated locomotion in dynamic environments, a three-layered control architecture (illustrated in Fig. \ref{fig:architecture}) is proposed:  \newline
\noindent 1) The first layer leverages forward reachable-set based CBFs to synthesize a safety-critical planner for coordination in dynamic environments using second-order models.
\newline 
\noindent 2) The second layer generates optimal ground reaction forces (GRFs) via SRB dynamics to track the safety-ensured trajectories from the first layer. 
\newline
\noindent 3) The third layer generates whole-body motion for each agent while imposing optimal GRFs and safety-ensured trajectories from the other two layers. This layer utilizes the full-order model of each agent with input-output (I-O) linearization \cite{khalil2002nonlinear} of virtual constraints to generate the desired torque inputs. 

The paper is organized as follows. Section \ref{sec:preliminaries} revisits the concept of forward reachable set and CBFs and extends it to the setting of safety-critical coordination control. Section \ref{sec:planner} formulates the safety-critical coordination planner for the system in dynamic environments. Section \ref{subsec:srbmpc} introduces the SRB dynamics with MPC, which designs the desired GRFs when tracking the safety-ensured velocities. Section \ref{subsec:qpvcctrl} introduces a whole-body motion control formulated as Quadratic Programming (QP) with I-O linearization technique.
In Section \ref{sec:results}, we discuss the experimental results of the proposed safety-critical coordination control in dynamic environments.  
We present conclusions in Section \ref{sec:conclusion}.

\section{Safety in Dynamic Environments}
\label{sec:preliminaries}

In this section, we will examine the safety-critical coordination of $N$ legged robots in dynamic environments using the framework of CBFs with forward reachable set. We consider a control-affine system for the $i$-th agent:
\begin{equation}\label{eq:controlaffinesystem}
    \dot{\bm{x}}_i = f_i(\bm{x}_i) + g_i(\bm{x}_i)\bm{u}_i,
\end{equation}
with state $\bm{x}_i(t)\in \mathcal{X}_i$
and control input $\bm{u}_i(t)\in \mathcal{U}_i$. Here $i\in \mathcal{I}:=\{1, \ldots, N\}$ represents the indexing for each individual agent.
We assume $f_i: \mathcal{X}_i \rightarrow \mathbb{R}^{n_i}$ and $g_i: \mathcal{X}_i\rightarrow\mathbb{R}^{n_i\times m_i}$ are Lipschitz continuous.

\begin{figure}[t!]
\centering
\includegraphics[draft=false, width=\linewidth]{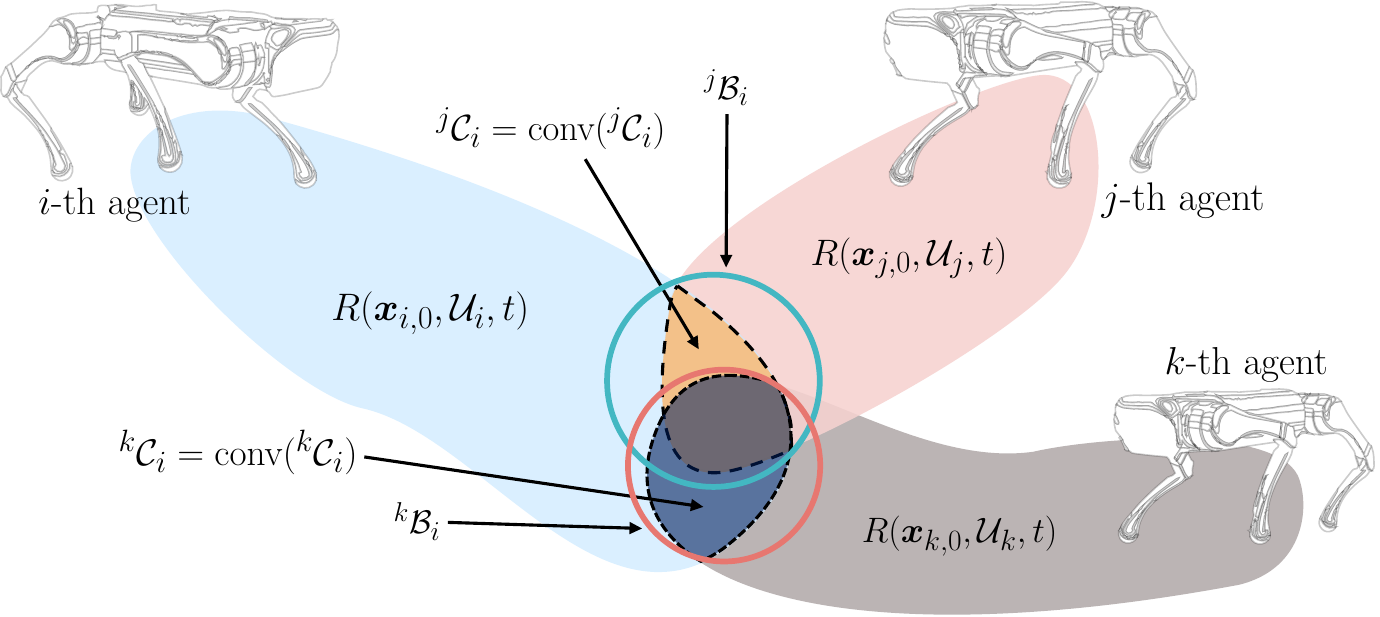}
\vspace{-2.0em}
\caption{Illustration of the forward reachable set of each agent and the region of potential collision.}
\label{fig:rpc}
\vspace{-1.5em}
\end{figure}

\subsection{Forward Reachable Set and Region of Potential Collision}
Consider a control-affine system in \eqref{eq:controlaffinesystem} evolving over a time horizon with some fixed time limit:

\gap
\begin{definition}\label{def:reachableset}
(\hspace{-0.01cm}\cite{lavalle2006planning})
    Let the \textit{forward reachable set} $R(\bm{x}_{i,0}, \mathcal{U}_i,t) \subset \mathcal{X}_i$ is the set of all states of the $i$-th agent that can be reached up to and including time $t$ when the initial state is $\bm{x}_{i,0}$ as:
    \begin{alignat}{2}
        R(\bm{x}_{i,0}, \mathcal{U}_i,t):=\{&\bm{x}_i\in\mathcal{X}_i: \exists \bm{u}_i\in\mathcal{U}_i \,\, \textnormal{and} && \nonumber \\
        &\exists\tau\in[0,t] \,\,\,\textnormal{s.t.}\,\,\, \bm{x}_i(\tau)=\bm{x}_i\}.
    \end{alignat}
\end{definition}
\gap

In the dynamic environment, forward reachable set can be defined on each movable object. To realize the area of possible collision within time $t$ for the legged robots while coordinating in dynamic environments,
\textit{region of potential collision} is defined as the intersections of forward reachable sets as described in Fig. \ref{fig:rpc}:

\gap
\begin{definition}
Let the \textit{region of potential collision} of the $i$-th agent in time $t$ with respect to the $j$-th object as:
\begin{equation}\label{eq:rpc}
    {^j}\mathcal{C}_{i} := \mathcal{P}_3(R(\bm{x}_{j,0}, \mathcal{U}_j,t)) \cap \mathcal{P}_3(R(\bm{x}_{i,0}, \mathcal{U}_i,t)),
\end{equation}
\gap
where $\mathcal{P}_3(R(\bm{x}_{i,0}, \mathcal{U}_i,t)):=\{\phi\in\mathbb{R}^3: \phi = p_3(\bm{x}_i)\}$ is a spatial reachable set of the $i$-th agent and $p_3: R(\bm{x}_{i,0}, \mathcal{U}_i,t_i) \rightarrow \mathbb{R}^3$ is a function to get a spatial position of the $i$-th agent. 
\end{definition}
\gap

Additionally, convex hull of ${^j}\mathcal{C}_i$ can be noted as $\textnormal{conv}({^j}\mathcal{C}_i)$ and the ball ${^j}\mathcal{B}_i \subset \mathbb{R}^2$ that includes the $\textnormal{conv}({^j}\mathcal{C}_i)$ can be defined as:
\begin{equation}\label{eq:rpcball}
{^j}\mathcal{C}_i \subseteq \textnormal{conv}({^j}\mathcal{C}_i) \subseteq {^j}\mathcal{B}_i:=\{\bm{z}\in \mathbb{R}^2 : \Vert \bm{z} - {^j}\bm{q}_i\Vert \leq {^j}r_i\},
\end{equation}
where $\Vert \cdot \Vert$ is 2-norm and ${^j}\bm{q}_i\in \textnormal{conv}({^j}\mathcal{C}_i)$ is the point that minimize the ${^j}r_i\in\mathbb{R}_{>0}$. Moreover, ${^j}\mathcal{B}_i$ specification in \eqref{eq:rpcball} can be represented with a tuple $({^j}\bm{q}_i,{^j}r_i)$.

Notably, the construction of the region of potential collision is not limited to a forward reachable set. 
In section \ref{sec:planner}, we will introduce the region of potential collision with a projected forward reachable set onto the direction of approaching to guarantee safety in dynamic environments.

\subsection{Safety of Individual Agent and Control Barrier Function}
We consider \eqref{eq:controlaffinesystem} to be safe if the system's state $\bm{x}_{i}(t)$ is located within a \textit{safe set} $\mathcal{S}_i \subset \mathcal{X}_i$ for all time, i.e., if $\mathcal{S}_i$ is forward invariant: 

\gap
\begin{definition}\label{def:safety}
System \eqref{eq:controlaffinesystem} is \textit{safe} with respect to the set $\mathcal{S}_i$ where the set $\mathcal{S}_i$ is \textit{forward invariant} along \eqref{eq:controlaffinesystem}: if $\forall\bm{x}_{i,0} \in \mathcal{S}_i$$\implies$$x_{i}(t)\in\mathcal{S}_i$, $\forall\bm{x}_{i}(0)=\bm{x}_{i,0}$ and $\forall t \geq 0$. 
\end{definition}
\gap

\begin{remark}(\textit{Safe set and region of potential collision})
    The determination of the safe set is based on the configuration of the system. For example, it can represent the relative distance between the $i$-th robot and ${^j}\mathcal{B}_i$ in \eqref{eq:rpcball} to prevent collisions in dynamic environments.
\end{remark}
\gap

Throughout the paper, we define the safe set $\mathcal{S}_i$ as the \textit{superlevel set} of a continuously differentiable function $h_{i}:\mathcal{X}\rightarrow \mathbb{R}$:
\begin{eqnarray}\label{eq:safeset}
    \mathcal{S}_i & := & \{ \bm{x}_i \in \mathcal{X}_i : h_{i}(\bm{x}) \geq 0 \}, \\
    \partial \mathcal{S}_i & := & \{ \bm{x}_i \in \mathcal{X}_i : h_{i}(\bm{x}) = 0 \}, 
\end{eqnarray}
where, $\bm{x}(t)\in \prod_{i \in \mathcal{I}}\mathcal{X}_i$.
CBFs can then be utilized as a tool to synthesize controllers that are provably safe:

\gap
\begin{definition}
    A continuously differentiable function $h_i:\mathcal{X}\rightarrow\mathbb{R}$ is a \textit{control barrier function (CBF)} for \eqref{eq:controlaffinesystem} on $\mathcal{S}_i$ if there exists an extended class $\mathcal{K}_\infty$ function $\alpha_i \in \mathcal{K}^e_\infty$ such that:
    \begin{equation}\label{eq:cbfdefinition}
        \sup_{u\in\mathcal{U}} \,\,\,\, \dot{h}_i(\bm{x},\bm{u}) \geq -\alpha_i(h_i(\bm{x}))
    \end{equation}
    holds  $\forall\bm{x} \in \mathcal{S}$, where control input $\bm{u}(t)\in \prod_{i \in \mathcal{I}} \mathcal{U}_i$ and 
    $$
    \dot{h}_i(\bm{x},\bm{u}) = \underbrace{\big(\nabla^{\top}_{\bm{x}}h_i(\bm{x})\big)f(\bm{x})}_{:= L_f h_i(\bm{x})} + 
    \underbrace{\big(\nabla^{\top}_{\bm{x}}h_i(\bm{x})\big)g(\bm{x})}_{:=L_g h_i(\bm{x})}\bm{u}
    $$
    with $L_f h_i(\bm{x})$ and $L_g h_i(\bm{x})$ the Lie derivatives \cite{khalil2002nonlinear} of $h_i$ with respect to $f(\bm{x})=\textnormal{col}(f_1(\bm{x}), f_2(\bm{x}), \cdots, f_N(\bm{x})) \in \mathbb{R}^n$ and $g(\bm{x})=\textnormal{diag}(g_1(\bm{x}), g_2(\bm{x}), \cdots, g_N(\bm{x})) \in \mathbb{R}^{n \times m}$, respectively.  Here $n=\sum_{i\in\mathcal{I}}n_i$, $m=\sum_{i\in\mathcal{I}}m_i$. Furthermore, ``$\col$'' and ``$\diag$'' denotes the column operator and matrix-to-block diagonal matrix operator, respectively.
\end{definition}
\gap 


\gap
\begin{theorem}[\hspace{-0.01cm}\cite{ames2016control}]\label{th:cbf}
\emph{If $h_i$ is a \textit{CBF} for \eqref{eq:controlaffinesystem}, then any locally Lipschitz continuous controller $\bm{u}=k_i(\bm{x})$ satisfying
\begin{equation}\label{eq:cbfsafecondition}
    \dot{h}_i(\bm{x}, k_i(\bm{x})) \geq -\alpha_i(h_i(\bm{x})), \qquad \forall \bm{x}_i\in \mathcal{S}_i
\end{equation}
guarantees that \eqref{eq:controlaffinesystem} is safe with respect to $\mathcal{S}_i$.}
\end{theorem}
\gap

Safety-critical control for the $i$-th agent to avoid unsafe set can be achieved by solving a QP with \eqref{eq:cbfsafecondition} as constraint when minimizing the difference between the desired and actual control input for the $i$-th agent:
\begin{alignat}{4}\label{eq:qpforone}
    k_i(\bm{x})& = &  \argmin_{\bm{u} \in \mathcal{U}}&\Vert\bm{u}-\bm{u}^{d}\Vert^{2}_{\mathbf{P}_s}\\
    &&\mathrm{s.t.} & \,\,\, L_f h_i(\bm{x}) + L_g h_i(\bm{x}) \bm{u} \geq -\alpha_i h_i(\bm{x}), \nonumber
\end{alignat}
where $\bm{u}^d$ is the desired control input for the agent, $\mathbf{P}_s$ is a positive definite matrix, $\Vert \bm{u} \Vert^{2}_{\mathbf{P}_s}=\bm{u}^{\top}\mathbf{P}_s \bm{u}$ and $\alpha_i$ is a positive number that can be tuned. The result of the QP is $k_i(\bm{x})$,  the (pointwise) optimal control input that meets the CBF constraint to ensure the safety of the $i$-th agent. The constraint on the decision variable, $\bm{u} \in \mathcal{U}$, implies that the optimal control input needs to take values in the admissible control input set.
The QP given in \eqref{eq:qpforone} is able to address the safety of individual agents in relation to the region of possible collision within their respective environments.



\section{Layered Controller for Safe Coordination}
\label{sec:planner}
The objective of this section is to present the layered control architecture for the quadrupedal robots' safe coordination. Here we note that the framework introduced in Section \ref{sec:preliminaries}
is applicable to a wide range of robotic systems in dynamic environments. However, we specialize its application here to the locomotion of quadrupedal robots that require safe coordination. To address the safety-critical coordination of these quadrupedal robots, we propose a control architecture comprised of three layers: (top layer) a safety-critical trajectory planner level with the second-order model, (middle layer) an optimal GRFs trajectory planner based on MPC with SRB dynamics, and (bottom layer) a real-time control level for the full-order dynamics leveraging virtual constraints with I-O linearization technique in QP formulation.
This three-level approach effectively tracks safe trajectories on legged robot platforms in dynamic environments.

\subsection{Safe Set with Region of Potential Collision}
The safety-critical trajectory planner employs second-order models as the representation of the system to guarantee the planned trajectory satisfies the safety constraints with respect to the region of potential collision. The second-order model for the $i$-th agent can be written as:
\begin{equation}\label{eq:secondordermodel}
    \begin{bmatrix}
        \dot{\bm{\varphi}}_i\\
        \ddot{\bm{\varphi}}_i
    \end{bmatrix} = 
    \begin{bmatrix}
        0 & \mathbf{I} \\ 0 & 0
    \end{bmatrix} \begin{bmatrix} \bm{\varphi}_i \\ \dot{\bm{\varphi}}_i \end{bmatrix} +
    \begin{bmatrix} 0\\ \mathbf{I} \end{bmatrix}\bm{\vartheta}_i,
\end{equation}
where $\mathbf{I}$ represents identity matrix, $\bm{\vartheta}_i\in\mathcal{V}_i\subset\mathbb{R}^2$ denotes the input of the second-order system, and $\bm{\varphi}_i\in\mathcal{D}_i\subset\mathbb{R}^2$ represents the planar positions of the $i$-th agent ($i\in\mathcal{I}$).

Utilizing the feedback linearizing control law, given by $\bm{\vartheta}_i(\bm{\varphi}_i, \dot{\bm{\varphi}}_i, \bm{\beta})=-\beta_1 (\bm{\varphi}_i - \bm{\varphi}_i^{\textnormal{des}}) - \beta_2 \dot{\bm{\varphi}}_i$, we can drive \eqref{eq:secondordermodel} toward the desired position, $\bm{\varphi}_i^{\textnormal{des}}$, where $\bm{\beta}=\textnormal{col}(\beta_1, \beta_2)$ and $\beta_1, \beta_2 \in \mathbb{R}_{>0}$. With the parameterized $\beta_1$ and $\beta_2$, the forward reachable set of \eqref{eq:secondordermodel} can be obtained as
\begin{alignat}{2}\label{eq:reachablesetforsecondorder}
    R(&\bm{\varphi}_{i,0}, \dot{\bm{\varphi}}_{i,0},t):=\{\bm{\varphi}_i\in\mathcal{D}_i: \exists \bm{\vartheta}_i(\bm{\varphi}_i, \dot{\bm{\varphi}}_i,\bm{\beta})\in\mathcal{V}_i \,\, , && \nonumber \\
    &\begin{bmatrix}\bm{\varphi}_i \\ \dot{\bm{\varphi}_i}\end{bmatrix} = \begin{bmatrix}\bm{\varphi}_{i,0} \\ \dot{\bm{\varphi}_{i,0}}\end{bmatrix} + \int_0^\tau \begin{bmatrix}\dot{\bm{\varphi}}_{i,0} + \int_0^\tau\bm{\vartheta}_i d\upsilon \\ \bm{\vartheta}_i(\bm{\varphi}_i, \dot{\bm{\varphi}}_i,\bm{\beta})\end{bmatrix} d\upsilon \,\, \textnormal{and} \nonumber\\
    & \,\, \tau\in[0,t] \,\,\,\textnormal{s.t.}\,\,\, \bm{\varphi}_i(\tau)=\bm{\varphi}_i\}.
\end{alignat}

Notably, \eqref{eq:reachablesetforsecondorder} assumes that the states and the control law of the object are known (e.g., legged agents). When the information of the moving object is limited with the current states, $(\bm{\varphi}_i, \dot{\bm{\varphi}}_i)$, without specified control law (e.g., moving obstacles or unidentified agents), its forward reachable set can be updated with
\begin{alignat}{2}\label{eq:reachablesetforobstacle}
    &R(\bm{\varphi}_{i,0}, \dot{\bm{\varphi}}_{i,0},t) :=\{\bm{\varphi}_i\in\mathcal{D}_i: -\dot{\bm{\varphi}}_{i,0} \leq \bm{\zeta}_i \leq \dot{\bm{\varphi}}_{i,0} \,\, , && \nonumber \\
    &\bm{\varphi}_i = \bm{\varphi}_{i,0} + \int_0^\tau \!\! \! \! \bm{\zeta}_i d\upsilon \,\, \textnormal{and} \,\, \tau\in[0,t] \,\,\,\textnormal{s.t.}\,\,\, \bm{\varphi}_i(\tau)=\bm{\varphi}_i\}.
\end{alignat}

To represent the proximity of the forward reachable set of $i$-th agent and $j$-th object in terms of the approaching direction, a set consisting of the normed elements in the projected forward reachable set of $i$-th agent onto $\bm{\varphi}_{ji}:=\bm{\varphi}_j-\bm{\varphi}_i/\| \bm{\varphi}_j-\bm{\varphi}_i\|$ can be defined as
\begin{equation}
    \tilde{R}_{\bm{\varphi}_{ji}}:= \{ \|\phi\| \in \mathbb{R}: \phi \in \textnormal{proj}_{\bm{\varphi}_{ji}}(\mathcal{P}_2(R(\bm{\varphi}_{i,0}, \dot{\bm{\varphi}}_{i,0},t)))\},
\end{equation}
where $\textnormal{proj}_{\bm{\varphi}_{ji}}(\cdot)$ is a vector projection operator onto $\bm{\varphi}_{ji}$ and $\mathcal{P}_2(\cdot)$ is a planar reachable set of the $i$-th agent analogous to \eqref{eq:rpc}.

\begin{figure}[t!]
\centering
\includegraphics[draft=false, width=\linewidth]{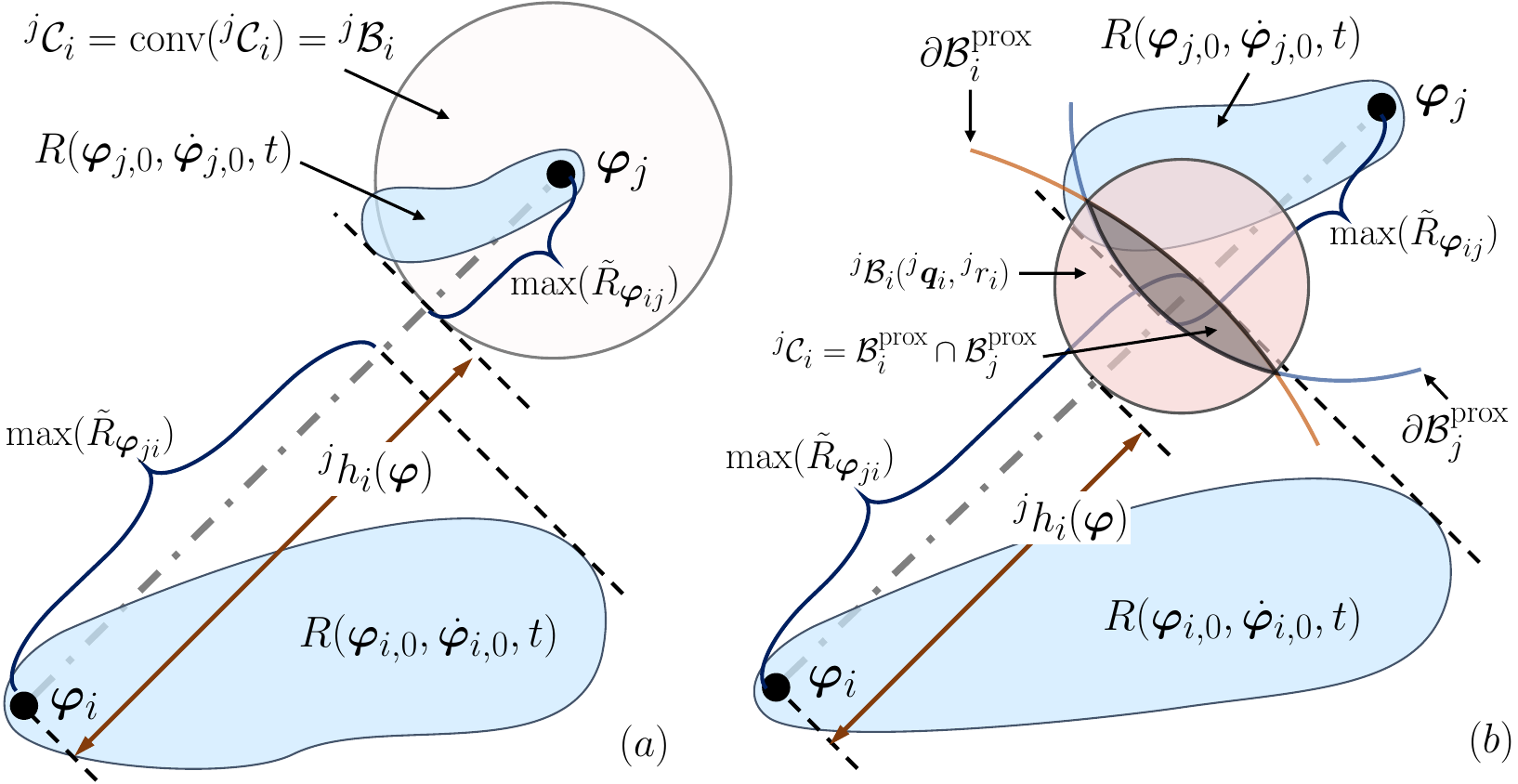}
\vspace{-2.0em}
\caption{Illustration of the safe set construction with second-order models (a) when the region of potential collision is an empty set and (b) when the region of potential collision is a non-empty set.}
\label{fig:frsimplementation}
\vspace{-2.0em}
\end{figure}

The proximity in dynamic environments can be categorized in two cases. The first case happens when $\textnormal{max}(\tilde{R}_{\bm{\varphi}_{ji}})+\textnormal{max}(\tilde{R}_{\bm{\varphi}_{ij}}) \leq \| \bm{\varphi}_j-\bm{\varphi}_i\|$, which is described in Fig. \ref{fig:frsimplementation} (a). In this case, the CBF can be defined as
\begin{equation}\label{eq:cbfforrpcball}
    {^j}h_i(\bm{\varphi}):=\Vert \bm{\varphi}_i-{^j}\bm{q}_i\Vert-{^j}r_i,
\end{equation}
where ${^j}\bm{q}_i = \bm{\varphi}_j$ and ${^j}r_i = \textnormal{max}(\tilde{R}_{\bm{\varphi}_{ij}})$, which also imply ${^j}\mathcal{C}_i = \textnormal{conv}({^j}\mathcal{C}_i) = {^j}\mathcal{B}_i$ when defining the region of potential collision.

The second case of proximity arises when $\textnormal{max}(\tilde{R}_{\bm{\varphi}_{ji}})+\textnormal{max}(\tilde{R}_{\bm{\varphi}_{ij}}) > \| \bm{\varphi}_j-\bm{\varphi}_i\|$, as depicted in Fig. \ref{fig:frsimplementation} (b). The region of potential collision can be expressed analogous to \eqref{eq:rpc} as
${^j}\mathcal{C}_{i} = \mathcal{B}_{i}^{\textnormal{prox}} \cap \mathcal{B}_{j}^{\textnormal{prox}}$,
where
\begin{equation}
\mathcal{B}_{i}^{\textnormal{prox}}:=\{\bm{z}\in \mathbb{R}^2 : \Vert \bm{z}-\bm{\varphi}_i\Vert \leq \textnormal{max}(\tilde{R}_{\bm{\varphi}_{ji}})\}.
\end{equation}
Moreover, ${^j}\mathcal{B}_{i} \supseteq {^j}\mathcal{C}_{i}$ can be articulated analogously to \eqref{eq:rpcball} by employing the tuple $({^j}\bm{q}_i,{^j}r_i)$, which especially incorporates parameters $\bm{\varphi}_i$, $\dot{\bm{\varphi}}_i$, $\bm{\varphi}_j$,  and $\dot{\bm{\varphi}}_j$. The CBF for the second case can be formulated similarly to \eqref{eq:cbfforrpcball}.

\subsection{Control Barrier Function for Second-order Model}\label{subsec:hocbf}
As the second-order model in \eqref{eq:secondordermodel} is employed, the control input, $\bm{\vartheta}_i$, is not included in the first-order derivative of CBF.
To formulate the control law in QP by using the CBF that includes the control input as one of its parameters, high-order CBF\cite{xiao2021high} is employed with a series of functions $\psi_i^0$, $\psi_i^1$, and $\psi_i^2$ in the form:
\begin{align}\label{eq:hocbf}
    \psi_i^0(\bm{\varphi})&:=h_i(\bm{\varphi}),\nonumber\\
    \psi_i^1(\bm{\varphi})&:=\dot{\psi}_i^0(\bm{\varphi}) + \alpha_i^1(\psi_i^0(\bm{\varphi})),\\
    \psi_i^2(\bm{\varphi}, \bm{\vartheta})&:=\dot{\psi}_i^1(\bm{\varphi}) + \alpha_i^2(\psi_i^1(\bm{\varphi})),\nonumber
\end{align}
where $\alpha^1_i, \alpha^2_i, \in \mathcal{K}^e_\infty$, $\bm{\varphi}=\textnormal{col}(\bm{\varphi}_i)$, $\bm{\vartheta}=\textnormal{col}(\bm{\vartheta}_i)$, and $i\in \mathcal{I}$. Associated with \eqref{eq:hocbf}, we further define a series of sets $\mathcal{S}^0_i$, $\mathcal{S}^1_i$, $\mathcal{S}^2_i$ in the form:
\begin{align}\label{eq:hosafeset}
    \mathcal{S}^0_i& := \{ \bm{\varphi}_i \in \mathcal{D}_i : \psi_i^0(\bm{\varphi}) \geq 0 \},\nonumber\\
    \mathcal{S}^1_i& := \{ \bm{\varphi}_i \in \mathcal{D}_i : \psi_i^1(\bm{\varphi}) \geq 0 \},\\
    \mathcal{S}^2_i& := \{ \bm{\varphi}_i \in \mathcal{D}_i : \psi_i^2(\bm{\varphi}, \bm{\vartheta}) \geq 0 \}.\nonumber
\end{align}
Notably, the number of functions and sets in the series is associated with the system model's relative degree\cite{khalil2002nonlinear}, while we need \eqref{eq:hocbf} and \eqref{eq:hosafeset} for guaranteeing the safety for the second-order model in \eqref{eq:secondordermodel}.

\gap
\begin{definition}
    A function $\psi_i^0(\bm{\varphi})$ is a \textit{high-order CBF (HOCBF)} for \eqref{eq:secondordermodel} if there exists and extended class $\mathcal{K}_\infty$ functions $\alpha_i^1$ and $\alpha_i^2$ such that
    \begin{equation}\label{eq:hocbfdef}
        \psi_i^2(\bm{\varphi}, \bm{\vartheta}) \geq 0, \,\,\, \forall \bm{\varphi}_i\in \mathcal{S}^0_i \cap \mathcal{S}^1_i \cap \mathcal{S}^2_i.
    \end{equation}
\end{definition}
\gap

\gap
\begin{theorem} (\hspace{-0.01cm}\cite{xiao2021high})
    If $\psi_i^0(\bm{\varphi})$ is a HOCBF for \eqref{eq:secondordermodel}, then any locally Lipschitz continuous controller $\bm{\vartheta} = k'_i(\bm{\varphi})\in\mathcal{V}$ satisfying
    \begin{equation}\label{eq:hocbfthm}
        \psi_i^2(\bm{\varphi}, k'_i(\bm{\varphi})) \geq 0, \,\,\, \forall \bm{\varphi}_i\in \mathcal{S}^0_i \cap \mathcal{S}^1_i \cap \mathcal{S}^2_i
    \end{equation}
    guarantess that the set $\mathcal{S}^0_i \cap \mathcal{S}^1_i \cap \mathcal{S}^2_i$ is forward invariant for \eqref{eq:secondordermodel} and \eqref{eq:secondordermodel} is safe with respect to $\mathcal{S}^0_i \cap \mathcal{S}^1_i \cap \mathcal{S}^2_i$.
\end{theorem}
\gap

We remark that, \eqref{eq:hocbf}-\eqref{eq:hocbfthm} associate with the environment with one object to be avoided. Analogous to \eqref{eq:cbfforrpcball}, left superscript can be used for describing HOCBFs for the dynamic environment with multiple objects to be avoided.
More specifically, multiple HOCBFs can be generated in the inequality constraint for the QP:
\begin{equation}
    L_f \Psi^1_i(\bm{\varphi}) + L_g \Psi^1_i(\bm{\varphi}) \bm{\vartheta} \geq -\mathbf{A} \Psi^1_i(\bm{\varphi}),
\end{equation}
where $\Psi^1_i(\bm{x}):= \textnormal{col}({^j}\psi^1_i)$, $j\in\mathcal{I}$, $j\!\!\neq\!\!i$, and $\mathbf{A}:=\textnormal{diag}({^j}\alpha^2_i)$. For the remainder of this paper, we will use the term CBF as shorthand for HOCBF for simplification.

\subsection{Planner for Safety-Critical Coordination Control}\label{subsec:hocbfplanner}
Safety-critical coordination control for \eqref{eq:secondordermodel} can be formulated with CBF defined in Section \ref{subsec:hocbf} as: 
\begin{alignat}{4}\label{eq:qpforagentstotal}
    &&\arg\!\min_{\bm{\vartheta}_s\in\mathcal{V}}&(\bm{\vartheta}_s-k_d(\bm \varphi))^\top \mathbf{P}_s (\bm{\vartheta}_s-k_d(\bm \varphi)) \nonumber \\
    &&\mathrm{s.t.} & \,\,\, L_f \Psi^1_i(\bm{\varphi}) + L_g \Psi^1_i(\bm{\varphi}) \bm{\vartheta}_s \geq -\mathbf{A} \Psi^1_i(\bm{\varphi})
\end{alignat}
where $\mathbf{P}_s$ is a positive definite matrix and $k_d(\bm \varphi)$ is a nominal desired control input without considering the safe coordination. Additionally, $\bm{\vartheta}_s$ is the safe control input of \eqref{eq:secondordermodel} with the consideration of CBFs. 
Here we note that $\Psi^1_i(\bm{\varphi})$ is composed of CBFs for the $i$-th agent with respect to the $j$-th object, ${^j}\psi^1_i(\bm{\varphi})$. Similarly, $\mathbf{A}=\diag({^j}\alpha^2_i)$ is composed of ${^j}\alpha^2_i$ for the $i$-th agent with respect to the $j$-th object.
Here we remark that, $\Psi^1_i(\bm{\varphi})$ also has ${^j}\alpha^1_i(\bm{\varphi})$ as implicit parameters associated with \eqref{eq:hocbf}.

\subsection{SRB dynamics Based Desired GRF Trajectory Planner}\label{subsec:srbmpc}

The middle-level GRF trajectory planner focuses on achieving safe coordination during locomotion by bridging the safe trajectories generated by the safety-critical planner and the whole-body motion controller, which utilizes nonlinear full-order dynamics. To bridge this, SRB dynamics are employed in the middle-level GRF trajectory planner to compute optimal GRFs that impose the full-order system to adhere to the safety-ensured trajectories.

In the notation of SRB dynamics, the state of the SRB model for the $i$-th agent can be written as
    $\bm{\varrho}_i:=\textnormal{col}(\bm{\sigma}_{i}, \dot{\bm{\sigma}}_{i}, R_i,\bm{\omega}_i^{B_i})\in\mathbb{R}^{18}$,
where the Cartesian coordinates of the center of mass (COM) of the $i$-th agent in the inertial frame, $\{O\}$, are written as $\bm{\sigma}_{i}\in \mathbb{R}^3$. The orientation of the body frame, $\{B_i\}$, of $i$-th agent with respect to the inertial frame, $\{O\}$ is represented by $R_i \in  \textnormal{SO}(3)$ and $\bm{\omega}_i^{B_i}\in\mathbb{R}^3$ denotes the angular velocity of $i$-th agent in the body frame, $\{B_i\}$. 
According to the principle of virtual work, the SRB dynamics for the $i$-th agent can be expressed as:
\begin{align}\label{eq:nonlinearSRBdyn}
    & \dot{\bm{\varrho}} = \textnormal{col}(
        \dot{\bm{\sigma}}_{i}\quad  
        \ddot{\bm{\sigma}}_{i}\quad  
        \dot{R}_i \quad 
        \dot{\bm{\omega}}_i^{B_i}) \\  
        & = \textnormal{col}(
    \dot{\bm{\sigma}}_{i} \,\,\,\,\, 
    \sum\frac{1}{m}\bm{f}_{\textnormal{GRF}}\!-\!\bm{g} \,\,\,\,\,
    R_{i}\!\!\begin{bmatrix}\bm{\omega}_{i}^{B_{i}}\end{bmatrix}_{\times} \,\,\,\,\,
    \Upsilon(\bm{\varrho}, \bm{f}_{\textnormal{GRF}}) \nonumber),
\end{align}  
where $\bm{\sigma}^{\textnormal{foot}}_{i}\in \mathbb{R}^3$ is the contact foot location with respect to $\{O\}$, $\bm{f}_{\textnormal{GRF}}$ denotes the ground reaction force on each contact foot, and the map $\begin{bmatrix} \cdot \end{bmatrix}_{\times}: \mathbb{R}^3 \rightarrow \mathfrak{so}(3)$ represents the skew-symmetric operator. Moreover, $I_i\in \mathbb{R}^{3\times 3}$ denotes the moment of inertia of the $i$-th agent with respect to its body frame and $\bm{g}$ represents the constant gravitational vector. More detailed system model can be found in \cite{kim2022layered}.

Analogous to the variational-based approach of \cite{chignoli2020variational, ding2021representation}, nonlinear SRB dynamics in \eqref{eq:nonlinearSRBdyn} can be linearized in the form,
    $\bm{\eta}_{k+t+1\vert t}\! =\! \mathbf{C}\vert_{\op} \bm{\eta}_{k+t\vert t} + \mathbf{D}\vert_{\op} \bm{f}_{\textnormal{GRF}, k+t\vert t} + \bm{\epsilon}_{\op},$
where $k\!=\!0,1,\cdots, N_\textnormal{h}\!-\!1$ and $N_{\textnormal{h}}$ is the control horizon. Here, $\bm{\eta}_{k+t\vert t}$ and $\bm{f}_{\textnormal{GRF},k+t\vert t}$ denotes the predicted states and inputs (GRFs) at time $k+t$ computed at time $t$, respectively.
Additionally, the matrices with subscript `$\op$', and $\bm{\epsilon}_\op$ is the system and input matrices for the linearized system and offset term calculated around $\bm{\eta}_t$ and $\bm{f}_{\textnormal{GRF},t-1}$.
Here we note that the $\bm{\eta}:=\textnormal{col}(\bm{\sigma}_{i}, \dot{\bm{\sigma}}_{i}, \bm{\xi}_i,\bm{\omega}_i^{B_i})\in\mathbb{R}^{12}$ is the modified state with  $\bm{\xi}_i\in\mathbb{R}^3$ that approximate $R_i \in  \textnormal{SO}(3)$ around the operation point $R^{\op}_i$ with $R_i = R^{\op}_i \textnormal{exp}(\begin{bmatrix}\bm{\xi}_i\end{bmatrix}_{\times})$.


The MPC algorithm with SRB dynamics can thus be written as the optimization problem:
\begin{alignat}{4}\label{eq:srbmpc}
&\quad \min_{\bm{\eta},\, f_{\textnormal{GRF}}\in \mathcal{FC}} &&\|\bm{\eta}_{t+N_h|t}-\bm{\eta}^{\des}_{t+N_h|t}\|_{\mathbf{P}}^{2} \,\, + \\
&&& \sum_{k=0}^{N_{\textnormal{h}}-1} (\|\bm{\eta}_{k+t|t} - \bm{\eta}^{\des}_{k+t|t}\|_{\mathbf{Q}}^{2}+ \|\bm{f}_{\textnormal{GRF}, k+t|t} \|_{\mathbf{R}}^{2}) \nonumber\\
&\quad\quad\textrm{s.t.}&&\,\, \bm{\eta}_{k+t+1\vert t}\! =\! \mathbf{C}\vert_{\op} \bm{\eta}_{k+t\vert t} + \mathbf{D}\vert_{\op} \bm{f}_{\textnormal{GRF}, k+t\vert t} + \bm{\epsilon}_{\op}, \nonumber
\end{alignat}
where $k=0,1,\cdots,N_{\textnormal{h}}-1$, $\mathbf{P}$, $\mathbf{Q}$, and $\mathbf{R}$ are positive definite matrices. 
Additionally, $\mathcal{FC}$ denotes the friction cone condition on each foot contacting the ground. Here we remark that the first element of the optimal state, $\eta^{\star}_{t+1\vert t}$ and optimal GRFs, $\bm{f}^{\star}_{\textnormal{GRF},t\vert t}$, computed from \eqref{eq:srbmpc} is applied to the whole body controller as desired trajectories.

\begin{figure}[t!]
\centering
\includegraphics[draft=false, width=\linewidth]{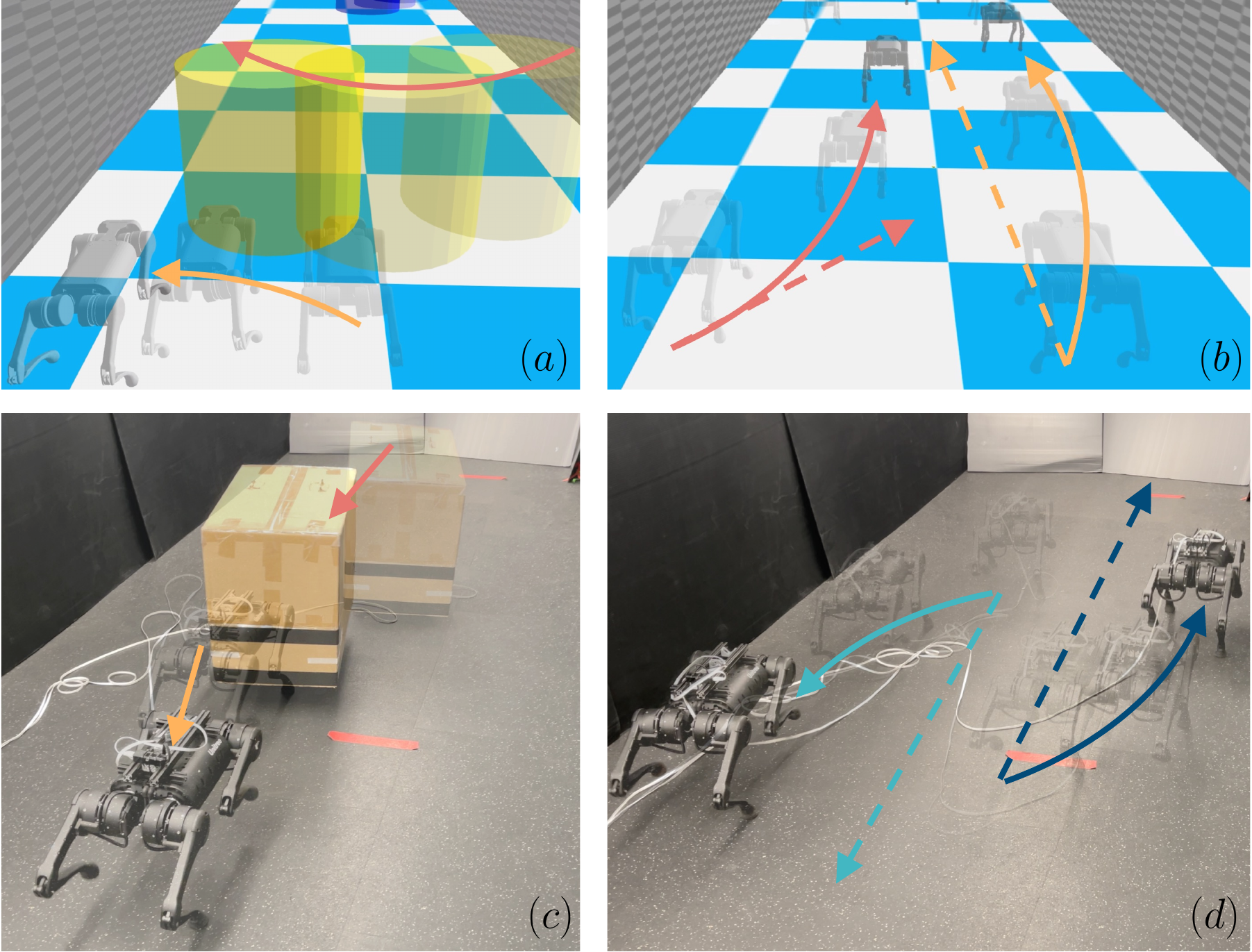}
\vspace{-2.0em}
\caption{Snapshots of the safety-ensured trajectory tracking (a) with moving obstacles in simulation, (b) with other agents in simulation, (c) with a moving obstacle in hardware experiments, and (d)  with the agents in hardware experiments.}
\label{fig:simandexp}
\vspace{-2.0em}
\end{figure}

\subsection{Virtual Constraint based Whole Body Controller} \label{subsec:qpvcctrl}

Given the safety-ensured trajectories and the optimal GRFs that impose the system to adhere to these trajectories, the whole-body controller is based on the virtual constraint controller \cite{hamed2020quadrupedal} in the task space to impose the full-order dynamics to track the prescribed trajectories and optimal GRFs.
The nonlinear full-order model of the $i$-th agent can be represented with generalized coordinates, $\bm{q}_i\in\mathcal{Q}\subset\mathbb{R}^{n_\textnormal{g}}$, where $\mathcal{Q}$ and $n_{\textnormal{g}}$ represent the configuration space and the dimension of generalized coordinates, respectively. Furthermore, $\bm{\tau}_i\in\mathcal{T}\subset\mathbb{R}^{n_\textnormal{j}}$ denotes the joint-level torques, where $\mathcal{T}$ is an admissible torques, and $n_\textnormal{j}$ is the number of joints. The overarching equations of motion for the nonlinear full-order model of the $i$-th agent can be represented by the Euler-Lagrange equation as follows:
\begin{equation}\label{eq:fullorderdynamics}
    \mathbf{D}(\bm{q}_i)\ddot{\bm{q}}_i + \mathbf{H}(\bm{q}_i,\dot{\bm{q}}_i) = \mathbf{B}\bm{\tau}_i+\sum \mathbf{J}_{\textnormal{c}}^{\top}(\bm{q}_i)\bm{f}_i,
\end{equation}
where $\mathbf{D}(\bm{q}_i)\in\mathbb{R}^{n_\textnormal{g} \times n_\textnormal{g}}$ and $\mathbf{H}(\bm{q}_i,\dot{\bm{q}}_i)\in\mathbb{R}^{n_\textnormal{g}}$ represent mass-inertia matrix and Coriolis, centrifugal and gravitational terms, respectively. Moreover, $\bm{f}_i$ and $\mathbf{J}_{\textnormal{c}}(\bm{q}_i)$ are the GRF on each ground contact point of $i$-th agent and the contact Jacobian matrices on $i$-th agent, respectively. Additionally, $\mathbf{B}\in\mathbb{R}^{n_\textnormal{g}\times n_\textnormal{j}}$ denotes the input distribution matrix. 

To impose the desired trajectories on the full-order model, the following time-varying output function for the $i$-th agent is defined:
\begin{equation}\label{eq:virtualconstraint}
    y_{i}(\bm{q}_{i},t):=\theta_{0}(\bm{q}_{i})-\theta^{\des}_{i}(t),
\end{equation}
which is termed \emph{virtual constraints}.
Here, $\theta_{0}(\bm{q}_{i})$ represents the set of holonomic quantites to be controlled, $\theta^{\des}_{i}(t)$ denotes the desired evolution of $\theta_{0}(\bm{q}_{i})$.
The variables to be controlled for each agent are chosen as the Euler angles, the Cartesian coordinates of the agent's COM, and the Cartesian coordinates of the swing feet in the inertial world frame. 
The desired evolution of the COM position and orientation is generated by the safety-critical planner in Section \ref{sec:planner}. Additionally, the desired swing foot trajectories are defined using a B\'ezier polynomial.
With the dynamics in \eqref{eq:fullorderdynamics}, these virtual constraints along the nonslippage condition at each stance foot are appended to a convex QP as follows \cite{hamed2020quadrupedal}:
\begin{alignat}{4}\label{eq:QPforvc}
&\min_{(\bm{\tau}_{i},\bm{f}_{i},\delta_i)}&& \,\,\,\frac{\nu_1}{2}\|\bm{\tau}_{i}\|^{2} + \frac{\nu_2}{2}\|\bm{f}_{i}-\bm{f}_i^{\textnormal{des}}\|^{2} + \frac{\nu_3}{2}\|\delta_i\|^{2}\\
&\quad\textrm{s.t.} && \textnormal{Output Dynamics with \eqref{eq:fullorderdynamics} and \eqref{eq:virtualconstraint}}\nonumber\\
& &&\textnormal{Nonslippage condition on each foot with \eqref{eq:fullorderdynamics}},\nonumber
\end{alignat}
where $\nu_1$, $\nu_2$, and $\nu_3$ are positive weighting factors, and $\bm{f}_i^{\textnormal{des}}$ represents the desired GRFs prescribed from the mid-level GRF planner. More detailed information about the derivation of output dynamics and nonslippage condition in \eqref{eq:QPforvc} can be found in \cite[Appendix A]{kim2022cooperative}. 
\section{Experimental Results}
\label{sec:results}
In this section, we provide the experimental results of the safety-critical coordination planner with the 18-degrees of freedom (DOFs) quadruped A1 developed by Unitree (see Fig. \ref{fig:simandexp}). The robot is modeled with the 6 DOFs accounting for the unactuated position and orientation of the body, while the remaining 12 DOFs correspond to the actuated leg joints. The weight of the robot is 12.45 (kg) and the center of the body is 0.26 (m) above the ground during locomotion. 
For experimental purposes, we use the laptop equipped with an Intel i7-1185G7 CPU operating at a frequency of 3.00 GHz with four cores and 16 GB of RAM. All QP-formulated optimization problems are solved using qpSWIFT \cite{pandala2019qpswift}.
In the experiments, each obstacle (see Figs. \ref{fig:simandexp}(a) and \ref{fig:simandexp}(c)) is kept at known states only, while each robot is kept at known states and control law (see Figs. \ref{fig:simandexp}(b) and \ref{fig:simandexp}(d)). When controlling the multiple robots for safety-critical coordination, the centralized method is employed to compute the safety-ensured trajectory of each robot.

\begin{figure}[t!]
\centering
\includegraphics[draft=false, width=\linewidth]{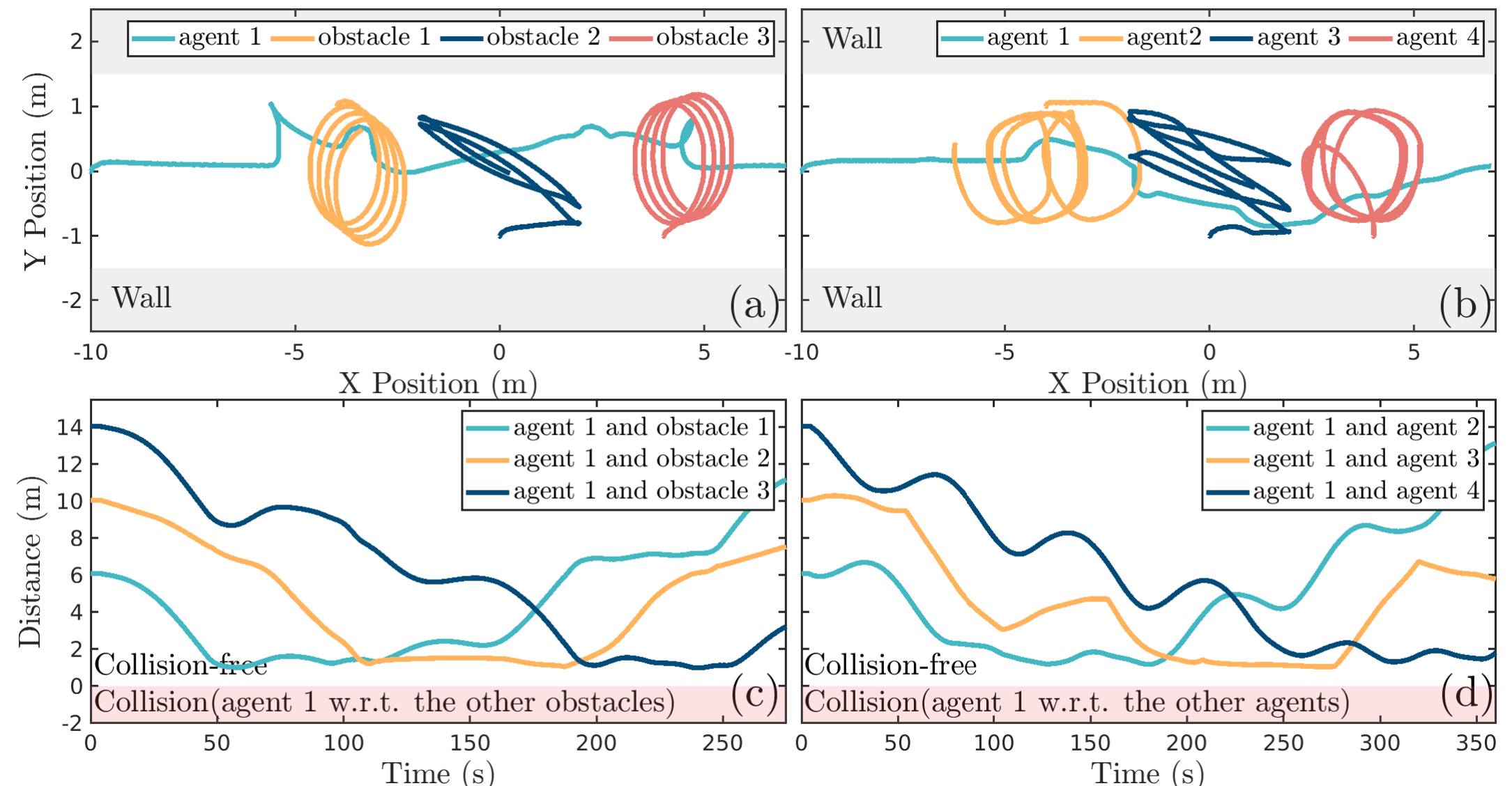}
\vspace{-2.0em}
\caption{Plots of the (a), (b) safety-ensured coordination trajectories and (c), (d) distance between objects in the simulation. The environment is composed of (a), (c) one agent and three obstacles and (b), (d) four agents.}
\label{fig:simtotal}
\vspace{-2.0em}
\end{figure}

\subsection{Simulation and Hardware Experiments}
We validate the effectiveness of the safety-critical coordination planner with full-order simulation experiments. We used RaiSim \cite{raisimpaper} for the simulation environment. 
We tested two simulation environments composed of one agent with three obstacles and four agents, respectively. The goal position of the target agent is $\textnormal{col}(10, 0)$ (m), while the desired moving pattern of three objects (i.e., three obstacles or three robots) is circular patterns with $1$ (m) radius with $\textnormal{col}(\pm 4, 0)$ (m) as a center and repeating a diagonal pattern between $\textnormal{col}(\pm 2, \mp 1)$ (m), respectively. $\mathbf{A}$ is $\textnormal{diag}(0.3\mathbf{I}_{3\times3}, 0.15\mathbf{I}_{2\times2})$ and $\textnormal{diag}(0.3\mathbf{I}_{6\times6}, 0.15\mathbf{I}_{8\times8})$ for each environment, respectively. Here, $0.15$ is for the CBF with respect to the left and right walls. Moreover, the corresponding value of $\alpha^1$ for $\alpha^2\!\!=\!0.3$ is $0.82$. 
The safety-critical planner is implemented as the top-level planner to ensure safe coordination that runs in $1$ (kHz).
In the middle level of the GRF trajectory planner, MPC with SRB dynamics runs in $500$ (Hz). 
The control horizon for the MPC is taken as $N_h \!\!=\! 5$. 
The stage cost gain is tuned as $\bm{Q} \!=\! \diag(3e5, 3e7,3e6,10^5\mathbf{I}_{3\times3},10^8\mathbf{I}_{3\times3},5\times10^3\mathbf{I}_{3\times3})$, the terminal cost gain is also tuned as $\bm{P} \!=\! 0.1\bm{Q}$, and the input gains are chosen as $\bm{R}_{f_{\textnormal{GRF}}} \!\!=\! 0.01\,\mathbf{I}_{12\times 12}$. 
Finally, the penalties in the whole-body controller to track the safety-ensured trajectory and optimal GRFs are chosen to be $\nu_1 \!=\! 10^2$, $\nu_2 \!=\! 10^3$, and $\nu_3 \!=\! 10^7$, resulting in stable locomotion.

The position evolutions of the full-order models in simulation experiments with obstacles and with other agents are described in Figs. \ref{fig:simtotal}(a) and \ref{fig:simtotal}(b), respectively. Here, the other obstacles keep their moving pattern while the agent tracks safety-ensured trajectory (see Fig. \ref{fig:simtotal}(a)), and all four agents maneuver with safety-ensured control law with modified trajectories for safe coordination (see Fig. \ref{fig:simtotal}(b)). The plots in Figs. \ref{fig:simtotal}(c) and \ref{fig:simtotal}(d) illustrate that the distance between the agent and obstacle or between agents, respectively, is maintained at a sufficiently safe margin, ensuring no collisions.


\begin{figure}[t!]
\centering
\includegraphics[draft=false, width=\linewidth]{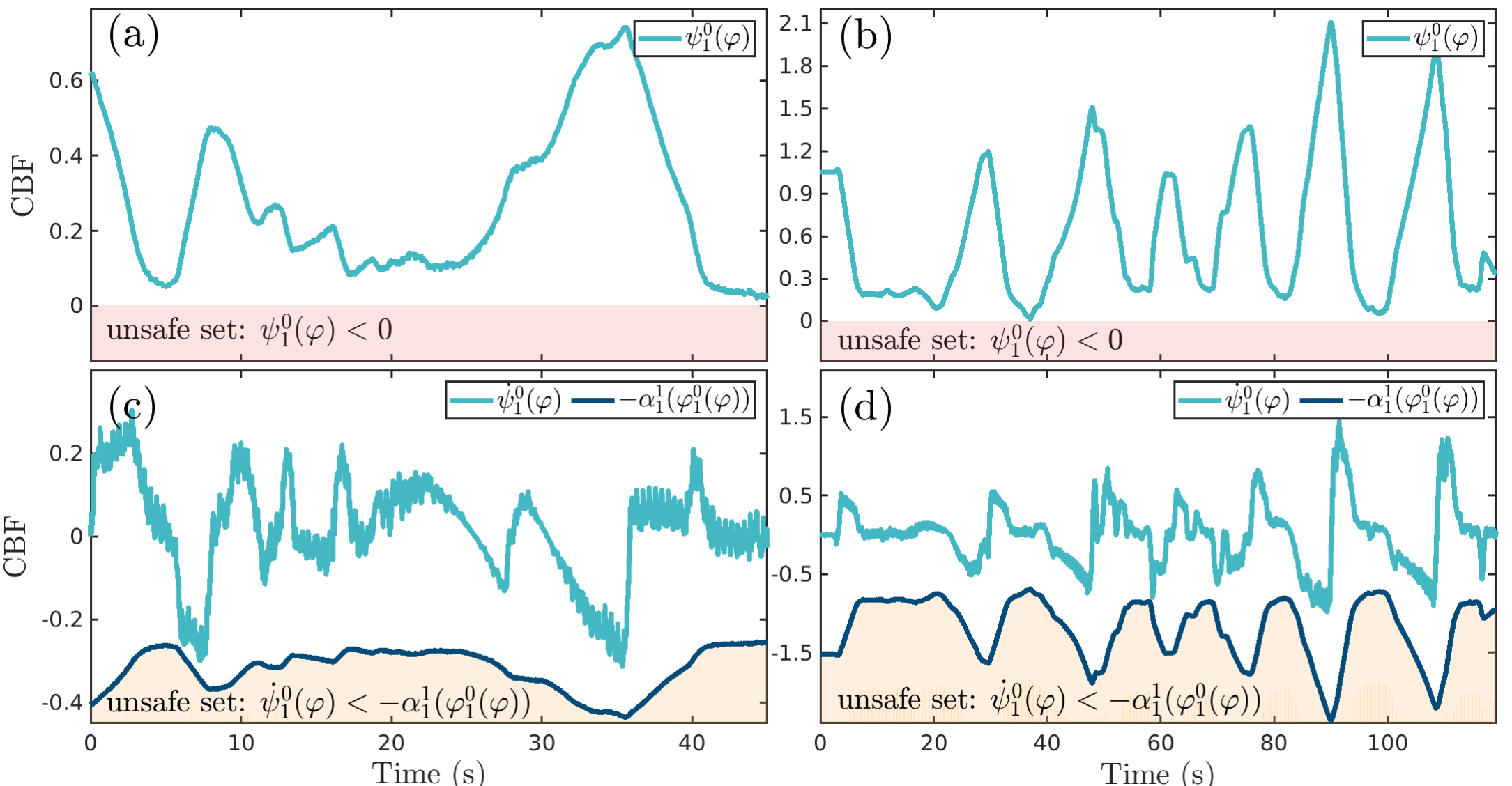}
\vspace{-2.0em}
\caption{Plots of the series of CBFs with hardware experiments. The environment is composed of (a), (c) one agent with one obstacle and (b), (d) two agents.}
\label{fig:hdwaretotal}
\vspace{-2.0em}
\end{figure}

We next study the performance of the closed-loop system with the safety-critical coordination controller on hardware. In hardware experiments, we tested two environments, one agent with an obstacle and two agents, respectively. We generated desired trajectories using the joystick. 
The gains and parameters used in hardware experiments are the same as those used in simulation experiments, while $\mathbf{A}$ is $\textnormal{diag}(0.3, 0.15\mathbf{I}_{2\times2})$ and $\textnormal{diag}(0.3, 0.15\mathbf{I}_{4\times4})$ for each environment setup in hardware experiments.
The CBF evolution in the environment with one agent and one obstacle is described in Figs. \ref{fig:hdwaretotal}(a) and \ref{fig:hdwaretotal}(c), while the CBF evolution in the environment composed of two A1 robots is described in Figs. \ref{fig:hdwaretotal}(b) and \ref{fig:hdwaretotal}(d). As the CBFs stay in the safe set area, we can conclude that the coordinated locomotion in dynamic environments tracks the safety-ensured trajectories. 

\section{Conclusion}
\label{sec:conclusion}

This work has demonstrated the potential of using CBFs to safely coordinate the legged robots in dynamic environments. Leveraging CBFs in the second-order model, we developed a locomotion controller with the safety-critical coordination planner that successfully accounted for the safety of legged robot locomotion in dynamic environments. Both simulation and hardware experiments showed the effectiveness of the proposed method in achieving safety-ensured coordination subject to various moving objects. In future work, we will explore how this framework could extend to more complex and uncertain dynamic environments. This includes the extension of the safety-critical coordination planner to three dimensions with perceptions to allow for varying terrain types.




\bibliographystyle{IEEEtran}
\bibliography{references}

\begin{thebibliography}{10}
\providecommand{\url}[1]{#1}
\csname url@samestyle\endcsname
\providecommand{\newblock}{\relax}
\providecommand{\bibinfo}[2]{#2}
\providecommand{\BIBentrySTDinterwordspacing}{\spaceskip=0pt\relax}
\providecommand{\BIBentryALTinterwordstretchfactor}{4}
\providecommand{\BIBentryALTinterwordspacing}{\spaceskip=\fontdimen2\font plus
\BIBentryALTinterwordstretchfactor\fontdimen3\font minus \fontdimen4\font\relax}
\providecommand{\BIBforeignlanguage}[2]{{%
\expandafter\ifx\csname l@#1\endcsname\relax
\typeout{** WARNING: IEEEtran.bst: No hyphenation pattern has been}%
\typeout{** loaded for the language `#1'. Using the pattern for}%
\typeout{** the default language instead.}%
\else
\language=\csname l@#1\endcsname
\fi
#2}}
\providecommand{\BIBdecl}{\relax}
\BIBdecl

\bibitem{fiorini1998motion}
P.~Fiorini and Z.~Shiller, ``Motion planning in dynamic environments using velocity obstacles,'' \emph{The international journal of robotics research}, vol.~17, no.~7, pp. 760--772, 1998.

\bibitem{schneider2003potential}
F.~E. Schneider and D.~Wildermuth, ``A potential field based approach to multi robot formation navigation,'' in \emph{IEEE International Conference on Robotics, Intelligent Systems and Signal Processing, 2003. Proceedings. 2003}, vol.~1.\hskip 1em plus 0.5em minus 0.4em\relax IEEE, 2003, pp. 680--685.

\bibitem{machado2016multi}
T.~Machado, T.~Malheiro, S.~Monteiro, W.~Erlhagen, and E.~Bicho, ``Multi-constrained joint transportation tasks by teams of autonomous mobile robots using a dynamical systems approach,'' in \emph{IEEE International Conference on Robotics and Automation}, 2016, pp. 3111--3117.

\bibitem{tagliabue2019robust}
A.~Tagliabue, M.~Kamel, R.~Siegwart, and J.~Nieto, ``Robust collaborative object transportation using multiple {MAVs},'' \emph{The International Journal of Robotics Research}, vol.~38, no.~9, pp. 1020--1044, 2019.

\bibitem{wehbeh2020distributed}
J.~Wehbeh, S.~Rahman, and I.~Sharf, ``Distributed model predictive control for {UAVs} collaborative payload transport,'' in \emph{IEEE/RSJ International Conference on Intelligent Robots and Systems}, 2020, pp. 11\,666--11\,672.

\bibitem{kim2022cooperative}
J.~Kim and K.~Akbari~Hamed, ``Cooperative locomotion via supervisory predictive control and distributed nonlinear controllers,'' \emph{Journal of Dynamic Systems, Measurement, and Control}, vol. 144, no.~3, 2022.

\bibitem{kim2022layered}
J.~Kim, R.~T. Fawcett, V.~R. Kamidi, A.~D. Ames, and K.~A. Hamed, ``Layered control for cooperative locomotion of two quadrupedal robots: Centralized and distributed approaches,'' \emph{arXiv preprint arXiv:2211.06913}, 2022.

\bibitem{fawcett2022distributed}
R.~T. Fawcett, L.~Amanzadeh, J.~Kim, A.~D. Ames, and K.~A. Hamed, ``Distributed data-driven predictive control for multi-agent collaborative legged locomotion,'' \emph{arXiv preprint arXiv:2211.06917}, 2022.

\bibitem{kim2023safety}
J.~Kim, J.~Lee, and A.~D. Ames, ``Safety-critical coordination for cooperative legged locomotion via control barrier functions,'' \emph{arXiv preprint arXiv:2303.13630}, 2023.

\bibitem{song2002potential}
P.~Song and V.~Kumar, ``A potential field based approach to multi-robot manipulation,'' in \emph{Proceedings 2002 IEEE International Conference on Robotics and Automation (Cat. No. 02CH37292)}, vol.~2.\hskip 1em plus 0.5em minus 0.4em\relax IEEE, 2002, pp. 1217--1222.

\bibitem{zhang2010dynamic}
M.~Zhang, Y.~Shen, Q.~Wang, and Y.~Wang, ``Dynamic artificial potential field based multi-robot formation control,'' in \emph{2010 IEEE Instrumentation \& Measurement Technology Conference Proceedings}.\hskip 1em plus 0.5em minus 0.4em\relax IEEE, 2010, pp. 1530--1534.

\bibitem{van2011reciprocal}
J.~Van Den~Berg, S.~J. Guy, M.~Lin, and D.~Manocha, ``Reciprocal n-body collision avoidance,'' in \emph{Robotics Research: The 14th International Symposium ISRR}.\hskip 1em plus 0.5em minus 0.4em\relax Springer, 2011, pp. 3--19.

\bibitem{guo2021vr}
K.~Guo, D.~Wang, T.~Fan, and J.~Pan, ``Vr-orca: Variable responsibility optimal reciprocal collision avoidance,'' \emph{IEEE Robotics and Automation Letters}, vol.~6, no.~3, pp. 4520--4527, 2021.

\bibitem{ames2016control}
A.~D. Ames, X.~Xu, J.~W. Grizzle, and P.~Tabuada, ``Control barrier function based quadratic programs for safety critical systems,'' \emph{IEEE Transactions on Automatic Control}, vol.~62, no.~8, pp. 3861--3876, 2016.

\bibitem{ames2019control}
A.~D. Ames, S.~Coogan, M.~Egerstedt, G.~Notomista, K.~Sreenath, and P.~Tabuada, ``Control barrier functions: Theory and applications,'' in \emph{2019 18th European control conference (ECC)}.\hskip 1em plus 0.5em minus 0.4em\relax IEEE, 2019, pp. 3420--3431.

\bibitem{molnar2021model}
T.~G. Molnar, R.~K. Cosner, A.~W. Singletary, W.~Ubellacker, and A.~D. Ames, ``Model-free safety-critical control for robotic systems,'' \emph{IEEE robotics and automation letters}, vol.~7, no.~2, pp. 944--951, 2021.

\bibitem{grandia2021multi}
R.~Grandia, A.~J. Taylor, A.~D. Ames, and M.~Hutter, ``Multi-layered safety for legged robots via control barrier functions and model predictive control,'' in \emph{2021 IEEE International Conference on Robotics and Automation (ICRA)}.\hskip 1em plus 0.5em minus 0.4em\relax IEEE, 2021, pp. 8352--8358.

\bibitem{lee2023hierarchical}
J.~Lee, J.~Kim, and A.~D. Ames, ``Hierarchical relaxation of safety-critical controllers: Mitigating contradictory safety conditions with application to quadruped robots,'' \emph{arXiv preprint arXiv:2305.03929}, 2023.

\bibitem{khazoom2022humanoid}
C.~Khazoom, D.~Gonzalez-Diaz, Y.~Ding, and S.~Kim, ``Humanoid self-collision avoidance using whole-body control with control barrier functions,'' in \emph{2022 IEEE-RAS 21st International Conference on Humanoid Robots (Humanoids)}.\hskip 1em plus 0.5em minus 0.4em\relax IEEE, 2022, pp. 558--565.

\bibitem{wang2017safety}
L.~Wang, A.~D. Ames, and M.~Egerstedt, ``Safety barrier certificates for collisions-free multirobot systems,'' \emph{IEEE Transactions on Robotics}, vol.~33, no.~3, pp. 661--674, 2017.

\bibitem{zhao2017defend}
S.~Zhao and Z.~Sun, ``Defend the practicality of single-integrator models in multi-robot coordination control,'' in \emph{2017 13th IEEE International Conference on Control \& Automation (ICCA)}.\hskip 1em plus 0.5em minus 0.4em\relax IEEE, 2017, pp. 666--671.

\bibitem{pickem2017robotarium}
D.~Pickem, P.~Glotfelter, L.~Wang, M.~Mote, A.~Ames, E.~Feron, and M.~Egerstedt, ``The robotarium: A remotely accessible swarm robotics research testbed,'' in \emph{2017 IEEE International Conference on Robotics and Automation (ICRA)}.\hskip 1em plus 0.5em minus 0.4em\relax IEEE, 2017, pp. 1699--1706.

\bibitem{chen2020guaranteed}
Y.~Chen, A.~Singletary, and A.~D. Ames, ``Guaranteed obstacle avoidance for multi-robot operations with limited actuation: A control barrier function approach,'' \emph{IEEE Control Systems Letters}, vol.~5, no.~1, pp. 127--132, 2020.

\bibitem{zhu2019chance}
H.~Zhu and J.~Alonso-Mora, ``Chance-constrained collision avoidance for mavs in dynamic environments,'' \emph{IEEE Robotics and Automation Letters}, vol.~4, no.~2, pp. 776--783, 2019.

\bibitem{lindqvist2020nonlinear}
B.~Lindqvist, S.~S. Mansouri, A.-a. Agha-mohammadi, and G.~Nikolakopoulos, ``Nonlinear mpc for collision avoidance and control of uavs with dynamic obstacles,'' \emph{IEEE robotics and automation letters}, vol.~5, no.~4, pp. 6001--6008, 2020.

\bibitem{zeng2021safety}
J.~Zeng, B.~Zhang, and K.~Sreenath, ``Safety-critical model predictive control with discrete-time control barrier function,'' in \emph{2021 American Control Conference (ACC)}.\hskip 1em plus 0.5em minus 0.4em\relax IEEE, 2021, pp. 3882--3889.

\bibitem{mitchell2005time}
I.~M. Mitchell, A.~M. Bayen, and C.~J. Tomlin, ``A time-dependent hamilton-jacobi formulation of reachable sets for continuous dynamic games,'' \emph{IEEE Transactions on automatic control}, vol.~50, no.~7, pp. 947--957, 2005.

\bibitem{bansal2017hamilton}
S.~Bansal, M.~Chen, S.~Herbert, and C.~J. Tomlin, ``Hamilton-jacobi reachability: A brief overview and recent advances,'' in \emph{2017 IEEE 56th Annual Conference on Decision and Control (CDC)}.\hskip 1em plus 0.5em minus 0.4em\relax IEEE, 2017, pp. 2242--2253.

\bibitem{lee2020reachability}
J.~Lee, J.~Ahn, E.~Bakolas, and L.~Sentis, ``Reachability-based trajectory optimization for robotic systems given sequences of rigid contacts,'' in \emph{2020 American Control Conference (ACC)}.\hskip 1em plus 0.5em minus 0.4em\relax IEEE, 2020, pp. 2158--2165.

\bibitem{khalil2002nonlinear}
H.~Khalil, \emph{Nonlinear systems}, 3rd~ed.\hskip 1em plus 0.5em minus 0.4em\relax Prentice Hall, 2002.

\bibitem{lavalle2006planning}
S.~M. LaValle, \emph{Planning algorithms}.\hskip 1em plus 0.5em minus 0.4em\relax Cambridge university press, 2006.

\bibitem{xiao2021high}
W.~Xiao and C.~Belta, ``High-order control barrier functions,'' \emph{IEEE Transactions on Automatic Control}, vol.~67, no.~7, pp. 3655--3662, 2021.

\bibitem{chignoli2020variational}
M.~Chignoli and P.~M. Wensing, ``Variational-based optimal control of underactuated balancing for dynamic quadrupeds,'' \emph{IEEE Access}, vol.~8, pp. 49\,785--49\,797, 2020.

\bibitem{ding2021representation}
Y.~Ding, A.~Pandala, C.~Li, Y.-H. Shin, and H.-W. Park, ``Representation-free model predictive control for dynamic motions in quadrupeds,'' \emph{IEEE Transactions on Robotics}, vol.~37, no.~4, pp. 1154--1171, 2021.

\bibitem{hamed2020quadrupedal}
K.~A. Hamed, J.~Kim, and A.~Pandala, ``Quadrupedal locomotion via event-based predictive control and qp-based virtual constraints,'' \emph{IEEE Robotics and Automation Letters}, vol.~5, no.~3, pp. 4463--4470, 2020.

\bibitem{pandala2019qpswift}
A.~G. Pandala, Y.~Ding, and H.-W. Park, ``qpswift: A real-time sparse quadratic program solver for robotic applications,'' \emph{IEEE Robotics and Automation Letters}, vol.~4, no.~4, pp. 3355--3362, 2019.

\bibitem{raisimpaper}
J.~Hwangbo, J.~Lee, and M.~Hutter, ``Per-contact iteration method for solving contact dynamics,'' \emph{IEEE Robotics and Automation Letters}, vol.~3, no.~2, pp. 895--902, 2018.

\end{thebibliography}

\end{document}